\documentclass{article}
\usepackage{PRIMEarxiv}
\usepackage[utf8]{inputenc} 
\usepackage[greek,english]{babel}
\usepackage[T1]{fontenc}    
\usepackage{hyperref}       
\usepackage{url}            
\usepackage{booktabs}       
\usepackage{amsfonts}       
\usepackage{nicefrac}       
\usepackage{microtype}      
\usepackage{lipsum}
\usepackage{fancyhdr}       
\usepackage{graphicx}       
\usepackage[title]{appendix}
\usepackage[square,numbers]{natbib}
\usepackage{float}
\usepackage{amsthm}
\usepackage{caption}
\usepackage[framemethod=tikz]{mdframed}
\usepackage{etoolbox} 
\usepackage{orcidlink}
\usepackage{listings}
\usepackage{tabularx}
\usepackage{multirow}
\usepackage{longtable}
\usepackage{booktabs}
\usepackage{makecell}
\usepackage{xcolor}
\usepackage{amsmath} 
\usepackage{enumitem}
\graphicspath{{media/}} 

\DeclareFontFamilySubstitution{LGR}{ptm}{artemisia}

\theoremstyle{definition}

\newmdtheoremenv[
  hidealllines=true,
  leftline=true,
  innerleftmargin=10pt,
  innerrightmargin=10pt,
  skipabove=10pt,
  skipbelow=10pt,
]{prompt}{Prompt}

\newcommand{\gk}[1]{\foreignlanguage{greek}{#1}}
\newcommand{\notS}{\ensuremath{\sim\!S}}
\newcounter{example}
\newcommand{\examplehead}[1]{\refstepcounter{example}\label{#1}\textbf{Example~\theexample:}}

\title{Semiotic Relations and Proof Methods: A Cross-Genre Study of Argument Structure with Large Language Models}

\author{
  Edirlei Soares de Lima \orcidlink{0000-0002-2617-3394}\\
  Academy for AI, Games and Media \\
  Breda University of Applied Sciences \\
  Breda, The Netherlands\\
  \texttt{soaresdelima.e@buas.nl} \\
  \And
  Marco A. Casanova \orcidlink{0000-0003-0765-9636} \\
  Department of Informatics \\
  PUC-Rio \\
  Rio de Janeiro, Brazil\\
  \texttt{casanova@inf.puc-rio.br} \\
  \And
  Antonio L. Furtado \orcidlink{0000-0003-3710-624X}\\
  Department of Informatics \\
  PUC-Rio \\
  Rio de Janeiro, Brazil\\
  \texttt{furtado@inf.puc-rio.br} \\
}

\begin{document}
\maketitle

\begin{abstract}
When a direct proof of a statement $S$ seems hard or even impossible to obtain, there may exist another statement (or set of statements) $S^{*}$, somehow related to $S$, on the basis of which $S$ can be proved. In order to investigate what options can be used to move from $S$ to $S^{*}$, four kinds of semiotic relations inspired by the four master tropes of semiotic research are briefly reviewed. Specifically, our syntagmatic, paradigmatic, antithetic and meronymic relations correspond, respectively, to metonymy, metaphor, irony and synecdoche. It is suggested that these four semiotic relations determine the options to move from $S$ to $S^{*}$, leading to proof by inference, proof by analogy, proof by contradiction, and proof by case analysis. To examine how the four relations are actually used across different kinds of argument, we complement the framework with an empirical study. We turn the four relations into explicit operational definitions and apply them to a cross-genre corpus of mathematical, legal, and everyday argument using a panel of large language models. We find that the relations are used very unevenly across genres: mathematical proofs draw on all four, whereas legal and everyday reasoning rely almost entirely on inference.
\end{abstract}


\section{Introduction}
\label{sec:intro}
 
The objective of this paper is to investigate what options one has to move from a statement $S$ that does not seem to be directly provable to some related statement (or set of statements) $S^{*}$, which could be shown to be true and to imply that $S$ itself must be true. We suggest that there are four ways to \emph{move} from $S$ to $S^{*}$, enabled by what we have categorized as \emph{semiotic relations}. These relations have been drawn from the so-called \emph{four master tropes}, a topic of major interest in the area of semiotic research \cite{Chandler2002}. It is no coincidence that `trope' comes from the Greek `\gk{τροπος}' from `\gk{τρεπειν}', `to turn', akin to the notion of \emph{moving} that underlies the present discussion.
 
Our four semiotic relations, together with their intuitive meaning, associated logical connectives, and corresponding tropes are shown in Table~\ref{tab:relations}.
 
\begin{table}[htbp]
\centering
\caption{Semiotic relations and their corresponding connectives and tropes.}
\label{tab:relations}
\begin{tabular}{@{}llll@{}}
\toprule
\textbf{Relation} & \textbf{Meaning} & \textbf{Connective} & \textbf{Trope} \\
\midrule
Syntagmatic  & Contiguity, sequence     & and     & Metonymy   \\
Paradigmatic & Similarity, alternatives & or      & Metaphor   \\
Antithetic   & Opposition, negation     & not     & Irony      \\
Meronymic    & Hierarchy, details       & part-of & Synecdoche \\
\bottomrule
\end{tabular}
\end{table}
 
These four tropes were characterized as fundamental, among the numerous rhetorical tropes popular in Greco-Roman antiquity \cite{Quintilian2001}, first in the XVI\textsuperscript{th} century \cite{Ramus2010} and again in the XVIII\textsuperscript{th} century \cite{Vico1968}. In modern times they were revived in a seminal study \cite{Burke1969}. Their universality has been repeatedly emphasized, with the indication that they may constitute ``a system, indeed \emph{the} system, by which the mind comes to grasp the world conceptually in language'' \cite{Culler1981}. Applications to several topics have been reported, for instance to worldviews and ideologies \cite{White1973} and, in our own work, to digital interactive composition of story-plots \cite{Lima2016, Lima2023, Lima2023sbgames, Lima2025b, Furtado2001}.
 
With respect to the names we assigned to the proposed semiotic relations, the terms `syntagmatic' and `paradigmatic' correspond to the two \emph{linguistic axes} of \citet{Saussure1995}. The term `antithetic' reflects the fact that, according to \citet{Burke1969}, the perspective induced by the irony trope is associated with \emph{dialectic}, which features \emph{antithesis} as a key concept expressing negation. Finally, in \citet{Winston1987}, wherein six types of part-of links are distinguished, one reads: ``We will refer to relationships that can be expressed with the term `part' in the above frames as `meronymic' relations after the Greek `meros' for part''.
 
Informally speaking, the preferred strategy to apply when $S$ is not directly provable is to look for other statements, in the same domain, from which $S$ could be deduced. If no clues are offered by the original domain, one may try to locate an analogue to $S$ in another domain, which may be more amenable to a successful treatment. Especially when $S$ is an assertion that something cannot hold, an often convenient option is to assume the contrary and then show that the assumption leads to an inconsistency. Finally, if a general proof of $S$ is unfeasible, one may break down the problem into an exhaustive list of cases, to be handled separately one by one. The main thrust of this paper is that these four options to prove a statement $S$ in connection with a statement (or set of statements) $S^{*}$ -- namely proof by inference, proof by analogy, proof by contradiction, and proof by case analysis -- are determined by the four semiotic relations mentioned before.
 
The framework is illustrated in Section~\ref{sec:applying} through a small set of carefully chosen examples. Whether the four relations it proposes describe argumentation as it actually occurs, once they are applied beyond such selected cases to large and varied collections of real arguments, is a question those examples cannot settle. The present paper takes it up empirically. We convert the four relations into explicit operational definitions, together with a residual category for arguments that none of them capture, and apply the resulting scheme to a cross-genre corpus of more than a thousand real arguments, drawn from established sources of mathematical, legal, and everyday reasoning, using a panel of large language models from independent families. Our aim is to characterize how the four relations are used across these genres, how consistently they can be applied, and how often an argument rests on a single relation or on several acting together, taking their grounding in the master tropes as given.
 
The paper is organized as follows. Section~\ref{sec:applying} explores the application of the four proof methods, relying on examples to illustrate the connection of each method with the respective enabling semiotic relation. Sections~\ref{sec:preliminary} and~\ref{sec:expressing} discuss a few problems arising from the complementary processes of finding a proof and expressing it convincingly. Section~\ref{sec:evaluation} reports the empirical study, in which the four relations are operationalized and applied across genres to characterize how they are used. Section~\ref{sec:concluding} contains concluding remarks.

\section{Applying the Semiotic Relations}
\label{sec:applying}
 
Certain statements are obviously true by definition, or are verifiable through a simple inspection. Direct proof that something exists merely requires exhibiting an instance, even though some work may be required to \emph{construct} it, as with the statement that there exist irrational numbers $a$ and $b$ such that $a^{b}$ is rational -- which is usually evidenced by producing some series of equalities (which, curiously, can only be checked symbolically since the first two cannot be computed over the domain $\mathbb{Q}$ of rational numbers):
\[
  a = \sqrt{2}, \qquad b = \log_{2} 9, \qquad a^{b} = 3
\]
but it often happens that no such direct proof is feasible.
 
To prove a statement $S$ in such circumstances, we can move to some other statement (or set of statements) $S^{*}$, which must be preliminarily shown to be linked to $S$ by a semiotic relation, and then try, recursively, to prove $S^{*}$. There are (at least) four such ``moves'', each of them corresponding to one of the rhetorical master tropes.
 
We say that a \emph{syntagmatic relation} holds between $S$ and $S^{*}$ if $S$ is a logical consequence of $S^{*}$. The associated trope is \emph{metonymy}. The resulting method is \emph{proof by inference}.
 
A \emph{paradigmatic relation} holds between $S$ and $S^{*}$ if after suitable mappings the relevant features of $S$ can be converted into features of $S^{*}$. The associated trope is \emph{metaphor}. The resulting method is \emph{proof by analogy}.
 
An \emph{antithetic relation} holds between $S$ and $S^{*}$ if $S^{*}$ could be shown to be inconsistent if \notS{} were true. The associated trope is \emph{irony}. The resulting method is \emph{proof by contradiction} (also called \emph{reductio ad absurdum}).
 
A \emph{meronymic relation} holds between $S$ and $S^{*}$ if $S^{*}$ is a set of statements into which $S$ can be decomposed exhaustively. The associated trope is \emph{synecdoche}. The resulting method is \emph{proof by case analysis}.
 
These proof methods are based on what might be called \emph{meta-theorems}, expressed below in a semiformal style (for a more rigorous treatment, see \cite{Enderton1972}), in terms of theories (denoted by $\Gamma$) and sentences (denoted by $\phi$ and $\gamma$):
 
\begin{description}[leftmargin=1.5em, style=nextline, font=\normalfont\itshape]
  \item[proof by inference] if $\Gamma \vdash (\gamma \rightarrow \phi)$ and $\Gamma \vdash \gamma$, \\ then $\Gamma \vdash \phi$.
  \item[proof by analogy] let $\pi[\Gamma] = \Gamma'$ be a faithful interpretation; \\ if $\Gamma' \vdash \phi'$ where $\phi' = \phi^{\pi}$, \\ then $\Gamma \vdash \phi$ \\ (noting that \emph{faithful} means $\gamma \in \Gamma \Leftrightarrow \gamma^{\pi} \in \Gamma'$).
  \item[proof by contradiction] if $(\Gamma ; \neg\phi)$ is inconsistent, \\ then $\Gamma \vdash \phi$ \\ (noting that \emph{inconsistent} means there is some $\gamma$ such that $(\Gamma ; \neg\phi) \vdash \gamma$ and $(\Gamma ; \neg\phi) \vdash \neg\gamma$).
  \item[proof by case analysis] let $\mu(\phi) = \{\phi_{1}, \phi_{2}, \ldots, \phi_{n}\}$ be an exhaustive decomposition; \\ if $\Gamma \vdash \phi_{i}$ for all $1 \leq i \leq n$, \\ then $\Gamma \vdash \phi$ \\ (noting that \emph{exhaustive} means $\{\phi_{1}, \phi_{2}, \ldots, \phi_{n}\}$ tautologically implies $\phi$).
\end{description}
 
\subsection{Proof by Inference}
 
\examplehead{ex:socrates} ``Socrates is mortal''. This time-honoured example recognizes that mortality is a human condition, as expressed by the rule: $\forall x\,(\mathit{human}(x) \rightarrow \mathit{mortal}(x))$. Since Socrates is known as a human being, the rule applies and the statement follows as a consequence.
 
\begin{figure}[htbp]
  \centering
  \includegraphics[width=0.6\textwidth]{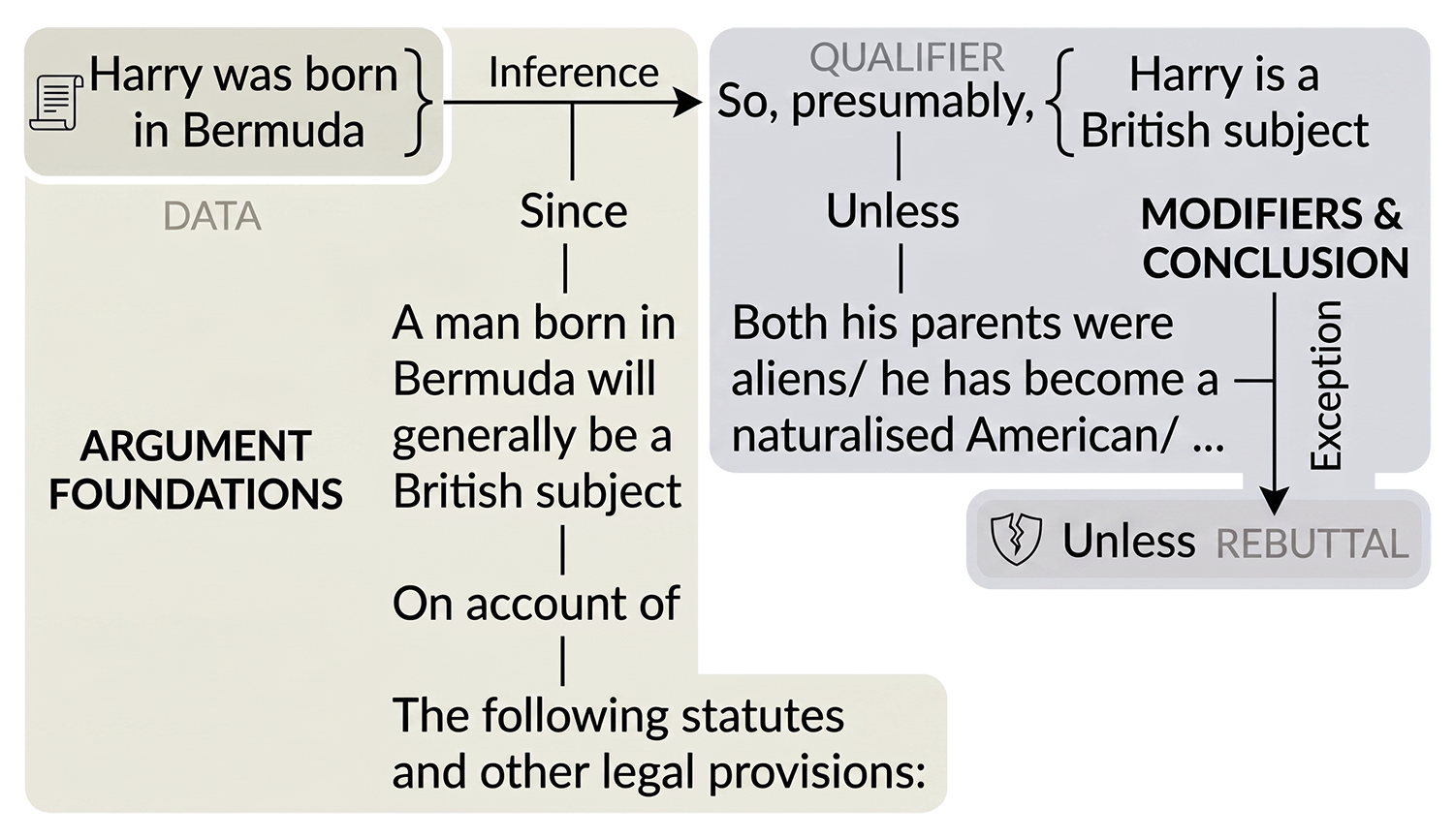}
  \caption{Toulmin's argument scheme \cite{Toulmin2003}.}
  \label{fig:toulmin}
\end{figure}

\medskip
\noindent\examplehead{ex:harry} ``Harry, who was born in Bermuda, is a British subject''. Stephen Toulmin has argued convincingly that the conventional syllogism structure must be expanded to deal with reasoning in the domain of justice \cite{Toulmin2003}. So it is not enough to consider what he calls the \emph{data} (Harry was born in Bermuda), the \emph{claim} (Harry is a British subject) and the \emph{warrant} (a man born in Bermuda is a British subject). To these three elements he adds a modality or, to use his own terms, a \emph{qualifier} (presumably), given that the rule admits exceptions that constitute a possible \emph{rebuttal} (unless both his parents were aliens, or he has become an American citizen, or \ldots). But, even more characteristic of legal argument, is the \emph{warrant} (statutes and other legal provisions); indeed the judicial system is governed by positive law (as opposed to natural law), which must have been officially established, and which may differ for different countries (e.g.\ notice among the exceptions the prevalence of \emph{ius sanguinis} over \emph{ius soli}, in contrast to Brazilian norms). Toulmin's scheme can best be comprehended under the form of a diagram, as shown in Figure~\ref{fig:toulmin}.
 
To Toulmin's remarks one must add that the existence of what he calls the `data' may not be recognized in justice if not officially registered as well (e.g.\ via a birth certificate). In database terminology this corresponds to the `closed world assumption' \cite{Casanova1987}. Also recall the assumption in criminal law that a defendant is judged `not guilty' (thus avoiding the term `innocent') unless proved responsible for the alleged offence, which in turn must have been exactly specified by a previous law (\emph{nullum crimen sine prævia lege pœnale}). All these considerations bring to mind the principle of `negation as finite failure', also explained in \cite{Casanova1987} -- \notS{} holds whenever $S$ neither resides in the database nor can be derived from the stored data and the rules that have been explicitly defined.

\subsection{Proof by Analogy}
 
\examplehead{ex:scheduling} ``There can be no efficient algorithm to determine the minimum number of schedules for tests of a group of students, so that no student will miss a test because its schedule coincides with that of some other course in which the student is enrolled''. Establishing non-conflicting schedules has an analogue in graph theory, if courses are mapped into nodes, and the fact that two courses $c_{1}$ and $c_{2}$ have one or more students in common is mapped into an edge connecting the nodes labelled $c_{1}$ and $c_{2}$. Then the original problem is converted into the problem of finding the chromatic number of a graph, which has been shown to be NP-complete (and hence of intractable computational complexity) \cite{Karp1972}.
 
\medskip
\noindent\examplehead{ex:monk} ``A Buddhist monk begins at dawn one day walking up a mountain, reaches the top at sunset, meditates at the top overnight until, at dawn, he begins to walk back to the foot of the mountain, which he reaches at sunset. Make no assumptions about his starting or stopping or about his pace during the trips. Is there a place on the path which the monk occupies at the same hour of the day on the two trips?'' The solution given in \cite{Turner1998} involves a close analogue for which, rather surprisingly, no mathematical treatment is required, and in fact the answer is immediately evident. The mappings involve \emph{blending} the scene of the monk climbing with that of his return. The action takes place in a single day, with the monk and his double walking in opposite directions -- and so inevitably meeting himself at some intermediate place.
 
\subsection{Proof by Contradiction}
 
\examplehead{ex:primes} ``There exists an infinity of prime numbers''. Assume, on the contrary, that the primes form a finite set $L = \{p_{1}, p_{2}, \ldots, p_{n}\}$. The proof dates from ancient times \cite{Euclid1956}. Taking all the primes in $L$, one can obtain: $P = p_{1} \times p_{2} \times \cdots \times p_{n} + 1$. The number $P$ calculated in this way is either a new prime, in which case we already have a contradiction, or a multiple decomposable into prime factors: $P = q_{1} \times q_{2} \times \cdots \times q_{m}$. But the $q_{i}$ should be different from the prime numbers in $L$, since $P$ is not divisible by any of them (the division would always yield 1 as remainder). So the $q_{i}$ would be new primes, again contradicting the \notS{} assumption.
 
\subsection{Proof by Case Analysis}
 
\examplehead{ex:absval} ``The absolute value of the sum of two non-zero numbers is less than or equal to the sum of their absolute values''. In formal notation: $|a + b| \leq |a| + |b|$. There seem to be four cases, which can be easily treated by elementary arithmetic:
 
\begin{quote}
case 1. if $a$ and $b$ are positive, the left side is equal to the right side;\\
case 2. if $a$ is positive and $b$ negative, the left side is less than the right side;\\
case 3. if $a$ is negative and $b$ positive, the left side is less than the right side;\\
case 4. if both $a$ and $b$ are negative, the left side is equal to the right side.
\end{quote}
 
Actually the four cases are reducible to three, by collapsing cases 2 and 3 in view of the commutative property of addition.
 
\medskip
\noindent\examplehead{ex:sum} ``The sum of all natural numbers from 0 to $n$ is equal to $n \times (n + 1) / 2$''. To show case by case that this holds for any value of $n$ would lead to an infinite process. Fortunately, thanks to a technique known as finite induction, the problem can be reduced to the following cases:
 
\begin{quote}
case 1. for $n = 0$, the result of computing the formula is 0, which is obviously correct;\\
case 2. assume that for $n = i$ the formula works correctly, i.e.: $0 + 1 + \cdots + i = i \times (i + 1) / 2$;\\
case 3. for $n = i + 1$, it must be shown that the formula yields $(i + 1) \times ((i + 1) + 1)/2$. This last case, called the induction step, can be established by using the assumption for $n = i$ and then performing a series of simple algebraic transformations:
$(0 + 1 + \cdots + i) + (i + 1) = i \times (i + 1) / 2 + (i + 1) =
(i \times (i + 1) + 2 \times (i + 1))/2 = (i + 1) \times (i + 2) / 2 =
(i + 1) \times ((i + 1) + 1)/2$.
\end{quote}
 
\medskip
\noindent\examplehead{ex:fourcolour} ``Four colours are enough to colour a geographical map so that no two adjacent political units have the same colour''. This is the so-called four colours conjecture, which defeated the attempts of many researchers for a long time, until being finally established as a proven theorem by two researchers working together in 1976 \cite{AppelHaken1976}. They first managed to identify an exhaustive list of cases, corresponding to 1936 ``irreducible configurations''. To handle such an overwhelming number of cases, they were forced to appeal to computer support. Subsequent efforts were made to reduce this number, but to our knowledge it still remains quite large.

\section{Finding a Proof}
\label{sec:preliminary}
 
Finding the proof of a statement $S$ and expressing the proof are more often than not two sharply different processes. In particular, to find a proof by inference, a person must start in a backward direction by applying a reasoning strategy called \emph{abduction} \cite{Peirce1998}. Its purpose is to search for some \emph{hypothesis} $S^{*}$ that may be used next to justify $S$. To perform abduction, one assumes that $S$ holds and then looks for some existing rule of the form $S^{*} \rightarrow S$ relating $S$ and $S^{*}$. In a sense, abduction involves traversing the rule in a right-to-left direction, inversely therefore to how we handle \emph{deduction}, on which the process of expressing a proof by inference is based. Recall that medical doctors rely on abduction while they try to trace back the observed symptoms to diseases that may have caused them, and that differential diagnosis becomes necessary if more than one disease is hypothesized.

Indeed, the reputed mathematician George Polya confirmed this primary opening role of abductive reasoning, when he asserted that guessing should precede proving \cite{Polya2014}: ``Finished mathematics presented in a finished form appears as purely demonstrative, consisting of proofs only. Yet mathematics in the making resembles any other human knowledge in the making. You have to guess a mathematical theorem before you prove it; you have to guess the idea of the proof before you carry out the details''. On the other hand, an assertion of the Indian mathematician Srinivasa Ramanujan \cite{Ranganathan1967}: ``Sir, an equation has no meaning for me unless it expresses a thought of GOD'', attributes his guesses to a sort of inspiration, which suggests that creative abduction may result from lucky intuition rather than systematic reasoning.
 
The rules themselves should have been formulated beforehand, typically by \emph{induction}, i.e.\ by observing that $S$ occurs whenever $S^{*}$ does, and that this can be attributed to logical implication or at the very least to probabilistic evidence, rather than to fortuitous coincidence (the \emph{post hoc ergo propter hoc} fallacy). After the advent of computers, \emph{data mining} runs \cite{Han2011} (involving statistical correlation and several other techniques) began to be routinely performed over large data repositories to discover such useful rules.
 
For proof by analogy, the preliminary search is even trickier. One must be able to look for analogues in domains other than that of the statement on hand, and abstract the essentials from knowledge expressed in a widely distinct formalism. Perhaps the required competence hinges on access to a repertoire of well-structured and well-indexed mental \emph{forms}, either characterized as ideas \cite{Plato1926}, or archetypes \cite{Jung1981}, or basic metaphors \cite{Lakoff2003}, or scripts \cite{Schank1977}, etc. Whether they are inborn or acquired is the topic of endless debate.
 
Children are encouraged very early in school to answer analogy questions in the form ``A is to B as C is to \emph{what}?''. Indeed proportionality is a helpful criterion to formulate the mappings between the features of the original statement and the candidate analogue. A modern discipline, \emph{case-based reasoning} \cite{Kolodner1993}, attempts to automate the search for analogues, ideally working on some rich computer-accessible library. One technique to construct such libraries involves extracting patterns from the observed detailed descriptions through \emph{most specific generalization} \cite{Ciarlini2009,Furtado1992}.
 
For proof by contradiction, determining $S^{*}$ can sometimes be almost immediate. In Example~\ref{ex:primes}, in opposition to the notion of an infinity of numbers with the property of being prime, one promptly perceives, without leaving the original domain, that a contrary notion is that the existing primes form a finite set -- and from that follows the idea of using the members of this set to construct the statement that will lead to a contradiction. But other problems are not so simple. We shall look at the famous Fermat's Last Theorem (proposed in 1637, just before his death), which was expressed by a simple algebraic equation, but was proved by contradiction much later \cite{Wiles1995}, using a rather advanced geometry result about the modularity of elliptic curves. So it combines analogy with contradiction (plus long series of inferences) and, on top of all that, it illustrates how crucial it is to \emph{restrict} the cases to be covered in a proof by case analysis to precisely what is required to prove the statement -- it became eventually clear that it suffices to consider \emph{semistable} elliptic curves. Once again as in Example~\ref{ex:fourcolour}, we shall only provide a very brief and very informal note, since a rigorous account would require mathematics well beyond the scope of this paper.
 
\medskip
\noindent\examplehead{ex:fermat} ``No three positive integers $a$, $b$, and $c$ can satisfy the equation $a^{n} + b^{n} = c^{n}$ for any integer value of $n$ greater than two''. Thanks to the effort of a number of researchers from 1637 to 1995 (when Wiles's paper was published), it was proved, case by case, that any solution to this deceptively simple equation could be used to generate a non-modular semistable elliptic curve, whereas it was also proved that all such elliptic curves had to be modular -- a contradiction that implies that there can be no solutions to the equation, thus finally transforming the conjecture into a theorem.

\section{Expressing a Proof}
\label{sec:expressing}
 
\begin{figure}[htbp]
  \centering
  \includegraphics[width=0.6\textwidth]{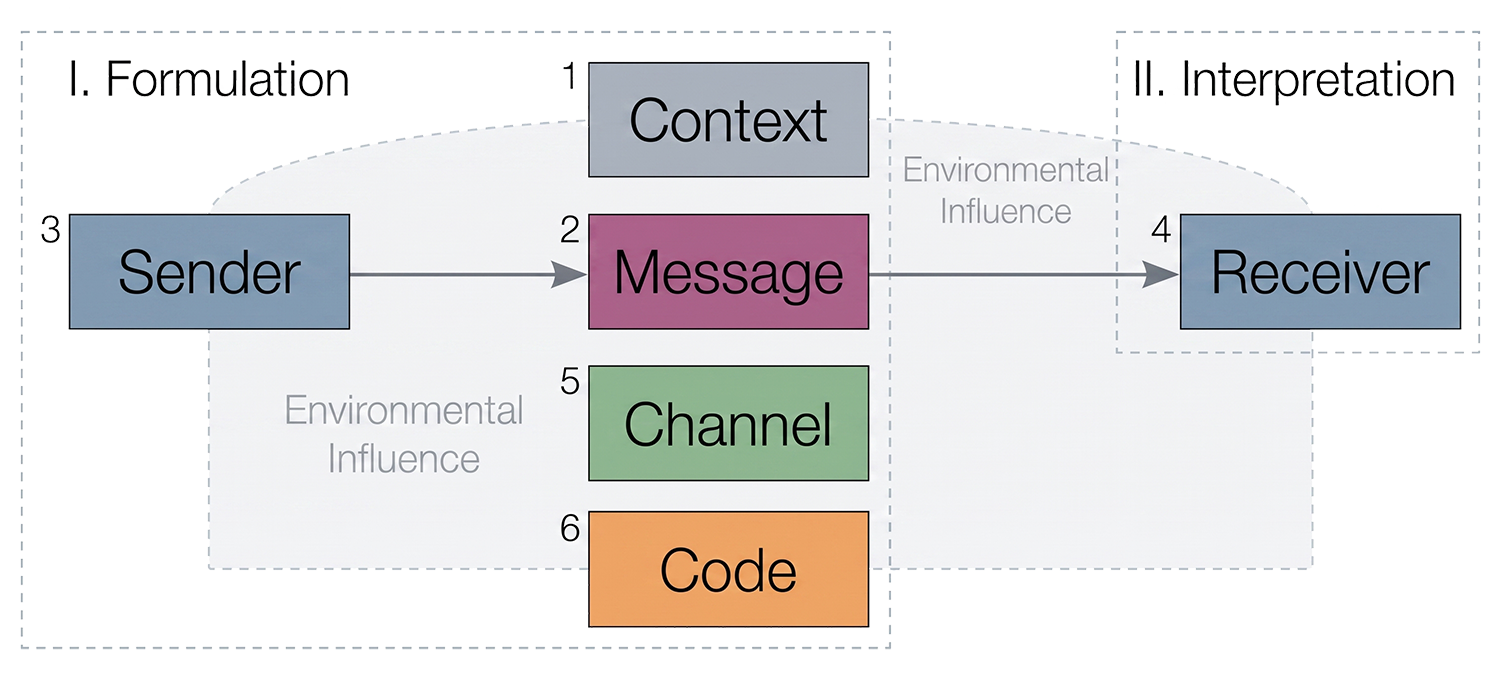}
  \caption{Jakobson's model of the communication process \cite{Jakobson1981}.}
  \label{fig:jakobson}
\end{figure}

Let us now turn to the second process mentioned at the beginning of the previous section, namely, having succeeded in proving that a sentence is true by a judicious application of methods such as those exemplified in Section~\ref{sec:applying}, how to suitably express the demonstration to other people. To realize what is involved it is convenient to view this as a \emph{communication process}, requiring our attention to at least the six items contained in the diagram shown in Figure~\ref{fig:jakobson}.
 
In words: the researcher (sender) who devised the proof formulates the demonstration (message) in some formal or informal language (code) and passes it through some medium (channel) to an interested person (receiver) who should be able to understand it. The cultural environment prevailing at a given place and time (context) imposes conditions that may exert a favourable or unfavourable influence on the outcome of the process. 

Of course the sender must make sure that the proof is correct with respect to both contents and form, but the effort can ultimately succeed only if the receiver can decode the demonstration to the point of actually learning it and taking maximum advantage of the new knowledge thus acquired. The choice of a formalism is sometimes crucial to this end. For instance, the use of finite induction for Example~\ref{ex:sum} above is considered by \citet{Chateaubriand2001} as inappropriate for teaching beginning students. Even if the algebraic manipulations can be followed by them, the stepwise argument would not ``relate meaningfully'' to the students, whereas a more effective presentation relying on a pictorial sketch would have a better chance of eliciting a reaction of ``dawning understanding''. \citet{Lakoff2000} provide several other similarly intuitive explanations using basic metaphors, for instance to show how complex numbers can be clearly understood by blending arithmetic and geometric notions.
 
More generally, the correct connection from sender to message in Jakobson's communication scheme is just one prerequisite of the process. It corresponds to the adequacy of the \emph{signifier} to the \emph{signified}, in Saussure's terms \cite{Saussure1995}. But communication must reach its final destination, the receiver, bringing to mind the three-element view -- \emph{object}, \emph{representamen}, \emph{interpretant} -- advocated by \citet{Peirce1998}. It is through this path that the human intellect can, although incompletely and imperfectly, grasp a glimpse of the real.
 
Perversely, even if something is understood correctly, a naive receiver may draw one or more wrong conclusions (cf.\ the notion of \emph{misconstruals} in \citet{Webber1986}) from it. It is a well known fact that statistical reports, though in themselves possibly correct, are very often misinterpreted out of inexperience or bad faith. But let us examine two kinds of wrong conclusions that may result from an undue application of a seemingly universal principle to a true statement: ``if a statement $S$ involving $a$ is true and $a = b$, the substitution of $b$ for $a$ yields a statement that is also true''.
 
First, take the true statement ``the expression $3 + 1 + 2$ contains three terms'', and note that $3 + 1 + 2 = 5 + 1$. By substitution, ``the expression $5 + 1$ contains three terms'' should be true, but it is patently false. Clearly the substitution could not have been done, since this particular statement is an argument \emph{de dicto}, whereas the value equality is a \emph{de re} consideration. Or we might say, perhaps, that the statement referred to a signifier and the comparison to a signified, in Saussure's terminology.
 
The second case is a little less trivial. Suppose the statement ``Gottlob believes that Venus is a planet'' is true, and consider the relatively well-known equality Venus $=$ Evening Star. The substitution, giving ``Gottlob believes that the Evening Star is a planet'' is not necessarily true, however. Even if Gottlob is aware of the equality, he may have never taken the trouble to perform the substitution, and therefore the maximum that we could say in this case, introducing a modality, is that ``Gottlob \emph{possibly} believes that the Evening Star is a planet''. The full-fledged substitution would only be warranted if both the equality and the substitution took place in Gottlob's head, i.e.\ at the level of Peirce's interpretant.

\section{An Empirical Evaluation of the Semiotic Relations}
\label{sec:evaluation}
 
Sections~\ref{sec:applying} through~\ref{sec:expressing} have developed the four relations and the two processes that surround any proof, its discovery and its communication. We now return to the relations themselves and ask a question that the illustrative examples of Section~\ref{sec:applying} cannot settle: whether they describe reasoning as it actually occurs across many arguments and different domains. To address this empirically, we provide operational definitions for the four relations and use a panel of large language models to apply them to a corpus of 1,126 arguments drawn from mathematics, law, and everyday reasoning.
 
Applying the relations at this scale enables an examination of questions left open by the examples in Section~\ref{sec:applying}. Specifically, we investigate the extent to which the four relations cover real argumentation, the uniformity of their usage, and the consistency with which a single argument is assigned a relation. We further determine whether the relations function as mutually exclusive alternatives or in combination, and, where the data permit, whether the labels reflect the structure of the argument rather than its specific lexical choices. The remainder of this section addresses these questions in turn.

\subsection{Methodology}
\label{sec:eval:methodology}

\subsubsection{Operational Definitions of the Relations}
\label{sec:eval:definitions}

The evaluation begins by converting the four relations into operational definitions applicable to any argument. Each definition opens with the wording from Section~\ref{sec:applying} and adds criteria specifying the classes of argument it covers. The four relation definitions below are inserted verbatim into the prompt supplied to the language models that perform the annotation (Section~\ref{sec:eval:instrument}), and are applied exactly as shown:

\begin{itemize}
  \item \textbf{SYN (syntagmatic; inference).} $S$ is a logical consequence of $S^{*}$. The reasoner locates a rule or prior result and applies it to reach $S$. Includes syllogism, rule application, chained deduction, and defeasible warrant-based argument.

  \item \textbf{PAR (paradigmatic; analogy).} After suitable mappings, the relevant features of $S$ are converted into features of $S^{*}$. The reasoner solves the mapped problem instead of the original one. The analogue often lies in another domain, as when a scheduling problem is mapped onto graph colouring, but it need not: mapping one case onto a structurally similar case within the same domain is still PAR, as when a problem is solved by blending a situation with a parallel version of itself. What matters is that a correspondence is drawn between two cases and the argument runs through it. Includes reduction, structural analogy, argument from a parallel case, and conceptual blending.

  \item \textbf{ANT (antithetic; contradiction).} $S^{*}$ is shown to be inconsistent under the assumption of \notS{}. Includes reductio ad absurdum and any argument whose force comes from the untenability of the denial. Concession is not ANT. Granting a point and then arguing past it (``Of course X, but Y''; ``Admittedly X; nevertheless Y'') is a rhetorical move, not an argument from inconsistency; label such passages by whatever establishes the main claim. Nor is mere contrast between two things ANT. The test is whether denying the claim is shown to lead to something untenable.

  \item \textbf{MER (meronymic; case analysis).} $S^{*}$ is a set of statements into which $S$ decomposes exhaustively, each handled separately. Includes case
  splits, exhaustion, and finite induction (base case + induction step).
\end{itemize}

The scheme also includes a residual category, NONE, designated for cases that satisfy none of the four relations or advance no argument, such as unsupported assertions or factual recitations without inference. This category records the absence of a relation rather than annotator uncertainty, and arguments that fit one relation but are difficult to place are assigned their best-fitting relation. Each argument is assigned a single \emph{dominant} relation, defined as the one upon which the argument depends such that its removal would leave the claim unestablished, alongside any number of \emph{subordinate} relations that perform genuine supporting work. The complete annotation prompt, including the verbatim wording for NONE and the model instructions, is presented in Appendix~\ref{app:prompt}.

\subsubsection{Corpus}
\label{sec:eval:corpus}
 
The evaluation is conducted on a corpus covering the three genres of reasoning illustrated in Section~\ref{sec:applying}, namely mathematical proofs, legal arguments, and everyday reasoning. The corpus contains 1,126 items, each between 50 and 300 words in length, drawn from three established sources and deduplicated. Its composition is summarized in Table~\ref{tab:corpus}, with the sources described below.
 
\begin{table}[htbp]
\centering
\caption{Composition of the corpus.}
\label{tab:corpus}
\begin{tabular}{@{}lllr@{}}
\toprule
\textbf{Genre} & \textbf{Source} & \textbf{Unit} & \textbf{N} \\
\midrule
Mathematics & NaturalProofs (ProofWiki) \cite{Welleck2021} & Theorem-proof pairs & 627 \\
Legal       & ECHR corpus \cite{Habernal2024}                     & Argument spans      & 400 \\
Everyday    & Microtext Corpus \cite{PeldszusStede2016}           & Single arguments    & 99  \\
\midrule
\textbf{Total} &                                              &                     & \textbf{1{,}126} \\
\bottomrule
\end{tabular}
\end{table}
 
The mathematical items comprise complete theorem-and-proof pairs drawn from a ProofWiki-derived collection \cite{Welleck2021}, with formatting markup removed. The everyday items consist of short single-argument texts on policy questions from the Argumentative Microtext Corpus \cite{PeldszusStede2016}. The legal items are argument spans from European Court of Human Rights decisions \cite{Habernal2024}, which also include expert annotations related to the argument type, a label that is withheld from the models and used only for the external validation presented in Section~\ref{sec:eval:external}.
 
The distinct argumentative structures of the three genres limit cross-genre comparison, as mathematical and everyday items constitute complete argumentative units while legal items function as components of a larger judgment. A legal item may state a rule, report a party's submission, or reach a conclusion supported by an adjacent part of the decision. Consequently, results are reported separately by genre rather than pooled. Since none of the three sources is a random sample of its genre, the reported distributions characterize these specific corpora rather than the genres in general.
 
The corpus, annotations, and analysis code are publicly available at \href{https://doi.org/10.5281/zenodo.22691375}{https://doi.org/10.5281/zenodo.22691375}.

\subsubsection{Annotation Procedure}
\label{sec:eval:instrument}
 
A panel of large language models applies the relations to the corpus, making labelling consistency measurable and ensuring that the models follow the operational definitions in Section~\ref{sec:eval:definitions} rather than relying on prior knowledge. This subsection details the panel, the labelling protocol, and the blinding conditions.

The panel comprises three large language models from independent families, as listed in Table~\ref{tab:models}, a design choice that reduces the risk that labels reflect the bias of a single type of model rather than the structure of the arguments. All three are open-weight models from unrelated developers, and the panel is intended to be diverse rather than matched in capability, since the study measures how consistently the scheme is applied rather than the accuracy of any one model. The models ran locally on a vLLM inference server,\footnote{https://vllm.ai/} with an OpenAI-compatible API, and were queried under identical settings.
 
\begin{table}[htbp]
\centering
\caption{The large language models comprising the annotation panel.}
\label{tab:models}
\begin{tabular}{@{}lll@{}}
\toprule
\textbf{Model} & \textbf{Developer} & \textbf{Parameters} \\
\midrule
Qwen3.8-27B  & Alibaba & 27B  \\
Gemma-4-31B  & Google  & 31B  \\
GLM-5.2-753B & Z.ai    & 753B \\
\bottomrule
\end{tabular}
\end{table}
 
Each model labelled every item three times at a sampling temperature of 0.7, producing nine independent judgements per item and 10,134 labels in total (1,126 items $\times$ 3 models $\times$ 3 runs). This repeated sampling at a non-zero temperature, rather than a single deterministic pass, allows the measurement of self-consistency, such that an item labelled identically across all three runs provides stronger evidence than one whose label varies. For each item, the model was first required to state, in free text, the claim being established ($S$), the support on which it rests ($S^{*}$), and how the two are connected, before recording the dominant relation, any subordinate relations, and a confidence rating on a three-point scale (1 = guessing, 2 = plausible, 3 = clear). Eliciting this statement prior to the label separates the identification of the argument from the assignment of a relation, allowing the two to be examined independently. The complete annotation prompt is provided in Appendix~\ref{app:prompt}.

The models were blinded to all information beyond the text of each item, which was presented in isolation with no indication of its genre, source, or selection procedure. The instruction wrapper named neither the originating work, nor the field of semiotics, nor the theorists associated with the four tropes. Furthermore, the four relation definitions were presented in a randomized order on each call, with NONE fixed in the final position, ensuring that the ordering of the options could not influence the labels. These measures concern only what the models see at inference time. Since the three corpus sources are publicly available, the item texts were most likely part of the models' pretraining data. This, however, does not contaminate the results in the way it would a benchmark, since the task provides no gold labels to recover and the scheme applied here is introduced in this paper. Prior exposure could at most shape how an item is read, and it is the consistency of that reading that Section~\ref{sec:eval:reliability} reports.

\subsection{Results}
\label{sec:eval:results}
 
The following subsections characterize the scheme's behaviour when applied to the corpus, covering the distribution of relations across genres, the manner in which those relations combine, the reliability with which they are applied, whether the prevalence of inference reflects genuine judgements rather than a default, and a bounded external test of validity.

\subsubsection{Distribution Across Genres}
\label{sec:eval:distribution}
 
The first result of the study to be analysed is the distribution of labels across the four relations, computed separately for each genre. Table~\ref{tab:distribution} reports the share of items, for each genre and category, whose label, defined as the most frequent of the nine judgements, falls within that category.
 
The distribution exhibits two patterns. First, the four relations cover argumentation broadly, with almost every item outside the legal genre belonging to one of them and the residual NONE category remaining small. Second, the relations are used very unevenly across genres. Mathematics draws on all four, where inference and case analysis account for most items (SYN 56.5\%, 95\%~CI $[52.6,60.3]$; MER 32.1\%, $[28.5,35.8]$) and contradiction occurs at a non-trivial rate (ANT 10.2\%, $[8.1,12.8]$). Legal and everyday reasoning, in contrast, concentrate heavily on inference (SYN 83.5\% and 93.9\%, respectively), with the other three relations nearly absent. The scheme thus covers reasoning broadly, though it distinguishes among the relations much more clearly within mathematics than in the other two genres.
 
\begin{table}[htbp]
\centering
\caption{Distribution of the modal label by genre, given as the count of items with the corresponding percentage of the $n$ items in each genre in parentheses.}
\label{tab:distribution}
\begin{tabular}{@{}lrrrrrr@{}}
\toprule
 &  & \multicolumn{5}{c}{\textbf{Modal Label}} \\
\cmidrule(lr){3-7}
\textbf{Genre} & \textbf{$n$} & \textbf{SYN} & \textbf{PAR} & \textbf{ANT} & \textbf{MER} & \textbf{NONE} \\
\midrule
Mathematics & 627 & 354 (56.5\%) & 5 (0.8\%)  & 64 (10.2\%) & 201 (32.1\%) & 3 (0.5\%)   \\
Legal       & 400 & 334 (83.5\%) & 18 (4.5\%) & 0 (0.0\%)   & 5 (1.2\%)    & 43 (10.8\%) \\
Everyday    & 99  & 93 (93.9\%)  & 5 (5.1\%)  & 1 (1.0\%)   & 0 (0.0\%)    & 0 (0.0\%)   \\
\bottomrule
\end{tabular}
\end{table}
 
The share of items labelled NONE differs significantly across genres. For genres composed of complete argumentative units, this share is minimal (mathematics 0.5\%, everyday 0.0\%), whereas the legal genre exhibits a higher rate (10.8\%, 95\%~CI $[8.1,14.2]$). This difference reflects the structure of the corpus described in Section~\ref{sec:eval:corpus} rather than a failure of the four relations to cover legal reasoning, as a legal item that reports a submission or states a rule without drawing a conclusion exhibits none of the four relations and is consequently labelled NONE.

\subsubsection{Combination and the Internal Structure of the Scheme}
\label{sec:eval:combination}
 
Beyond the marginal distribution, the labels reveal how the four relations interact. The share of items with a subordinate relation varies by genre, comprising 36.0\% in mathematics, 6.1\% in everyday, and 2.2\% in legal, based on a majority of votes. These numbers indicate that the relations co-occur rather than acting as exclusive alternatives, most often in mathematical proofs.

Broken down by the dominant relation, the combinations exhibit a clear asymmetry (Table~\ref{tab:combination}). Inference is nearly always self-sufficient, with only 5.9\% of items whose dominant relation is SYN carrying a subordinate relation. Contradiction and case analysis operate in the opposite direction, incorporating a subordinate relation in 80.0\% and 68.6\% of cases, respectively. In most of these instances, the subordinate relation is inference. Given that contradiction and case analysis appear almost exclusively in mathematics, this pattern is largely a feature of mathematical proofs. Arguments based on contradiction or exhaustive case splits typically perform inferential steps within the argument, whereas inference, when dominant, usually stands alone.

\begin{table}[htbp]
\centering
\caption{Combination structure by dominant relation, over the items whose modal label is one of the four relations. ``With subordinate'' gives the number and percentage of such items that carry at least one subordinate relation on a majority-of-votes basis.}
\label{tab:combination}
\begin{tabular}{@{}lrrl@{}}
\toprule
\textbf{Dominant} & \textbf{$n$} & \textbf{With subordinate} & \textbf{Most common subordinate} \\
\midrule
SYN & 782 & 46 (5.9\%)   & MER \\
PAR & 27  & 3 (11.1\%)   & SYN \\
ANT & 65  & 52 (80.0\%)  & SYN \\
MER & 204 & 140 (68.6\%) & SYN \\
\bottomrule
\end{tabular}
\end{table}

\subsubsection{Reliability}
\label{sec:eval:reliability}
 
Since the labels derive from an automated panel, their reliability depends on the consistency with which the scheme is applied, encompassing both within-model stability across repeated runs and agreement between independent models. Table~\ref{tab:reliability} reports three measures of this consistency. Self-consistency, defined as the rate at which a single model returns the same label for an item across three runs, ranges from 81.1\% to 89.1\% across the three models. Cross-model unanimity, the rate at which all three models agree on their per-model modal label, is 79.2\% (95\%~CI $[76.8,81.5]$), while chance-corrected agreement, measured by Krippendorff's $\alpha$ \cite{Krippendorff2004} over all nine judgements per item, is 0.712 overall.
 
\begin{table}[htbp]
\centering
\caption{Reliability of the annotation, with Krippendorff's $\alpha$ reported by
genre. Self-consistency and cross-model unanimity are corpus-wide.}
\label{tab:reliability}
\begin{tabular}{@{}lr@{}}
\toprule
\textbf{Measure} & \textbf{Value} \\
\midrule
Self-consistency across runs (range over models) & 81.1--89.1\% \\
Cross-model unanimity (all three agree)          & 79.2\% \\
Krippendorff's $\alpha$, overall                 & 0.712 \\
\quad mathematics                                & 0.748 \\
\quad everyday                                   & 0.582 \\
\quad legal                                      & 0.551 \\
\bottomrule
\end{tabular}
\end{table}

The reported per-genre Krippendorff's $\alpha$ values require careful interpretation. In genres where a single relation dominates, such as inference in the legal and everyday genres, expected chance agreement is high, which depresses $\alpha$ even when raw agreement remains identical. Consequently, relying on $\alpha$ alone would mischaracterize the more concentrated genres as less reliably measured, despite raw agreement being highest in precisely those contexts. The most frequent disagreement occurs between MER and SYN, followed by NONE against SYN and PAR against SYN, a pattern consistent with inference serving as the default with which the marked relations are most easily confused. These measures reflect the consistency of scheme application rather than the correctness of the labels.
 
\subsubsection{Confidence in the Inference Label}
\label{sec:eval:probe}
 
The distribution results presented in Section~\ref{sec:eval:distribution} point to the high frequency of inference, a pattern that holds across all genres and covers the large majority of legal and everyday reasoning. A natural objection is that models default to SYN when no other relation stands out, making its prevalence an artefact of the instrument rather than a property of the arguments. This objection leads to a testable prediction. Since a fallback is a choice made under uncertainty, SYN labels should show low confidence. The confidence rating recorded with each label allows us to test this prediction directly.
 
The confidence ratings contradict this prediction (Table~\ref{tab:confidence}). SYN labels exhibit a mean confidence of 2.93 on the three-point scale, with 93.4\% assigned the maximum rating. This value is comparable to case analysis (2.94) and contradiction (2.98), and exceeds that of analogy (2.78). Only analogy and the residual category NONE attract lower confidence, indicating that the models place their uncertainty there. Inference is therefore not the low-confidence fallback that a defaulting instrument would produce; the models are as sure of a SYN label as of a label from any of the marked relations.
 
\begin{table}[htbp]
\centering
\caption{Confidence by dominant label, computed over all judgements assigned each label. Mean confidence is on the three-point scale (1 = guessing, 2 = plausible, 3 = clear).}
\label{tab:confidence}
\begin{tabular}{@{}lrr@{}}
\toprule
\textbf{Label} & \textbf{Mean confidence} & \textbf{\% at maximum} \\
\midrule
SYN  & 2.93 & 93.4 \\
PAR  & 2.78 & 78.5 \\
ANT  & 2.98 & 97.6 \\
MER  & 2.94 & 94.1 \\
NONE & 2.76 & 75.7 \\
\bottomrule
\end{tabular}
\end{table}
 
\subsubsection{External Validity: a Bounded Discrimination Test}
\label{sec:eval:external}
 
The evaluations presented in the previous sections are fully based on the annotations produced by the large language models, which let us analyse how the relations are distributed and combined, how consistently they are applied, and whether inference was a default option instead of a real inference label. However, none of these measures compares labels against an independent standard, meaning they cannot demonstrate that a label reflects the structure of an argument rather than its surface wording. The legal corpus permits one bounded test of this specific distinction, employing the argument-type annotations from \cite{Habernal2024} as an expert key that is never presented to the large language models.
 
The expert scheme and our labels are not directly comparable, as one records the kind of legal argument while the other records the relation between $S$ and $S^{*}$. This is therefore a discrimination test, particularly because several of Habernal's argument types produce items that cite many earlier court decisions and appear similar on the surface, even though the underlying arguments differ. The central question is whether our labels can distinguish these types, with the clearest contrast lying between two such categories. In \emph{distinguishing a prior case}, a cited precedent is shown not to apply, so the argument proceeds through a comparison between cases, a structure our definitions predict as PAR. In \emph{prior case law}, earlier decisions are cited as authority for a rule that is then applied, which the definitions predict as SYN. Because both types cite prior decisions with comparable frequency, any difference in our labels between them cannot stem from the citations; rather, a higher rate of PAR labels on distinguishing than on prior case law would reflect the argument itself. The test also covers \emph{comparative law}, which argues from other jurisdictions (predicted PAR), and \emph{subsumption}, which applies a rule to the facts (predicted SYN). This mapping was fixed in advance, derived from the definitions and established before any labels were inspected.
 
Distinguishing and comparative law annotations are rare, and the 400 legal items in the main analysis contain too few to estimate a PAR rate. To address this, we add 40 further legal items of rare argument types from the same corpus \cite{Habernal2024}, selected on the expert argument type assigned by the original annotators and thus independently of any model label. Of these, the distinguishing and comparative law items are the ones this test uses. These items are used only for this test and are excluded from the distribution results in Section~\ref{sec:eval:distribution}.
 
The results indicate that the labels separate the types as predicted (Table~\ref{tab:external}). PAR labels appear on distinguishing and comparative law at a rate of 40.7\% (11 of 27, 95\%~CI $[24.5,59.3]$), compared to 5.1\% (4 of 79, $[2.0,12.3]$) on prior case law (Newcombe 95\%~CI $[17.9,54.5]$ \cite{Newcombe1998}; one-sided Fisher exact $p = 3.2\times 10^{-5}$). The separation is even stronger on distinguishing alone (50.0\%, 11 of 22; $p = 3.7\times 10^{-6}$), and the comparative-law items (5 in total) receive no PAR labels, so the effect is due entirely to distinguishing. Subsumption attracts SYN at 78.6\% (173 of 220, 95\%~CI $[72.8,83.5]$), well above the 45.5\% rate on distinguishing. Since SYN is the dominant label in the corpus, this serves as a sanity check rather than strong evidence.
 
\begin{table}[htbp]
\centering
\caption{Discrimination-test rates on legal items. Each row gives the rate of one relation within one expert type (or pair of types).}
\label{tab:external}
\begin{tabular}{@{}lrr@{}}
\toprule
\textbf{Measure} & \textbf{$n$} & \textbf{Rate} \\
\midrule
PAR on distinguishing $+$ comparative law & 27  & 40.7\% \\
PAR on distinguishing only                & 22  & 50.0\% \\
PAR on prior case law (baseline)          & 79  & 5.1\%  \\
SYN on subsumption                        & 220 & 78.6\% \\
\bottomrule
\end{tabular}
\end{table}

\section{Related Work}
\label{sec:related}
 
The idea that the search for a proof follows a small number of recurring methods has a long history in the study of mathematical practice. P\'olya's account of heuristics catalogues strategies such as reasoning by analogy, generalization, decomposition into subproblems, and working backwards from the goal \cite{Polya1945, Polya1954}, several of which correspond closely to the relations examined here. Lakatos, in turn, portrays mathematics as advancing through cycles of proof and refutation, in which a conjecture and its proof are reshaped by counterexamples \cite{Lakatos1976}. While these works describe how proofs are found and revised, the framework presented in this paper makes a different and more specific claim, that the choice among such methods is governed by the same four relations that underlie the master tropes.
 
The structure of argument more broadly has been formalized in ways that are also relevant to the framework. Toulmin analyses an argument into data, claim, warrant, and further qualifying elements \cite{Toulmin2003}, a model that appears in Example~\ref{ex:harry}; Walton's argumentation schemes catalogue the stereotypical patterns of presumptive reasoning, each paired with a set of critical questions \cite{Walton1996, WaltonReedMacagno2008}. Such schemes aim at fine-grained coverage of argumentative moves and are assembled from the bottom up, whereas the present framework derives four relations from the master tropes, at a higher level of abstraction. A single one of our relations subsumes many individual schemes, so the two are complementary rather than competing.
 
The empirical evaluation presented in Section~\ref{sec:evaluation}, by contrast, connects to the computational study of argument. It is most closely related to work in argument mining, which segments texts into argumentative units and classifies their roles or types. General-purpose schemes typically reduce an argument to premises and a claim linked by relations of support or opposition, whereas domain-specific efforts adopt richer typologies. For example, the corpus of European Court of Human Rights decisions used in our study annotates a fine-grained set of expert argument types \cite{Habernal2024}. Our relations operate at a different level. Rather than labelling components or cataloguing genre-specific patterns, they characterize the single move that carries an argument from its claim $S$ to the supporting statement $S^{*}$, and they are the same four across every genre.
 
The relations are also distinct from the discourse relations of frameworks such as Rhetorical Structure Theory \cite{MannThompson1988} and the Penn Discourse Treebank \cite{Prasad2008}, which annotate coherence links between adjacent spans of text, among them contingency, comparison and elaboration. These describe how a text is organized, not how a claim is established, and a passage may realize a discourse relation while making no argumentative move at all. The two kinds of annotation are largely orthogonal, which is why an existing discourse-relation corpus does not provide a ready-made comparison for the scheme we studied in this paper.
 
Alongside these annotation efforts, another line of work, surveyed by \citet{Li2024}, has been applying large language models to mathematical reasoning and proof directly, whether by generating natural-language proofs~\cite{Welleck2022}, translating them into the languages of formal proof assistants such as Lean~\cite{deMoura2021}, or searching for derivations~\cite{Polu2020}. Most ambitiously, the same techniques have begun to tackle open conjectures that had long resisted proof~\cite{Ke2026, Tzachristas2026}. Our aim in this work is different. We use models not to find or verify a proof but to label the relation that organizes a proof already given, and our mathematical items are complete natural-language proofs rather than formal derivations \cite{Welleck2021}. The analogy relation also connects more specifically to an active debate over whether language models reason analogically in a human-like manner \cite{Webb2023} or instead lean on surface-level correspondences \cite{LewisMitchell2025}. That debate concerns the models' own competence, whereas here analogy is one of the relations the models are asked to recognize in an argument they did not produce.
 
Because the labels analysed here are produced by large language models, the study also relates to a growing body of work that uses large language models as annotators \cite{Lima2025, Lima2026}. Such studies report that these models can match or exceed crowd workers on a range of classification tasks \cite{Tornberg2023}, though others caution that agreement among models is not evidence of correctness and that replacing human coders requires explicit evaluation \cite{Calderon2025}, and that model judgements can carry systematic biases of their own \cite{vanBlerck2025}. We adopt this instrument for its scale and reproducibility, and throughout we read its agreement as a measure of how consistently the scheme can be applied rather than of whether the labels are correct. More broadly, characterizing a theoretical typology at scale with language models, by operationalizing it as an explicit set of definitions, is a strategy we have applied to other conceptual schemes as well, including Northrop Frye's theory of fundamental genres \cite{Lima2026frye} and the investigative methods of fictional detectives \cite{Lima2025, Lima2026}.

\section{Concluding Remarks}
\label{sec:concluding}
 
This paper began from a simple observation: when a statement $S$ resists a direct proof, a proof can often be obtained by establishing a related statement $S^{*}$ instead. Around this observation we organized four semiotic relations connecting $S$ to $S^{*}$, corresponding to the four master tropes and to proof by inference, analogy, contradiction, and case analysis (Section~\ref{sec:applying}); we then considered the two processes that surround any proof, its discovery and its communication (Sections~\ref{sec:preliminary} and~\ref{sec:expressing}); and we characterized empirically how the four relations are used when applied to argumentation at scale (Section~\ref{sec:evaluation}). The empirical characterization shows that the relations cover argumentation broadly but are exercised very unevenly. Mathematical proofs draw on all four relations, whereas legal and everyday reasoning rely almost entirely on inference. Their internal structure is asymmetric, with inference largely self-sufficient while contradiction and case analysis typically recruit a subordinate relation, most often inference. A bounded external test in the legal genre, using expert argument-type labels withheld from the models, supports the view that the labels track the structure of an argument rather than its surface vocabulary.
 
The framework nevertheless remains a simplification, as any account of a complex practice must be. The full complexity of mathematical practice lies well beyond it: theorems such as the four-colour theorem (Example~\ref{ex:fourcolour}) and Fermat's Last Theorem (Example~\ref{ex:fermat}) still come in very lengthy proofs that demand proficiency across several domains, so much so that one is often compelled to accept such results on the authority of a few specialists. Efforts to convey them to a wider audience are only partly successful; \citet{Faltings1995}, presenting the ideas behind Fermat's Last Theorem, notes having passed over details judged to be of little interest to the nonspecialist.
 
Several limitations bound these findings. The most important is that every label analysed in our evaluation experiment is produced by large language models, so the agreement we observe measures how consistently the scheme can be applied, not whether the labels are correct as models trained on overlapping data can agree and still be wrong together. The external test offsets this only in part, since it reaches SYN and PAR in the legal genre alone and speaks to neither MER nor ANT. And because each genre is drawn from a single corpus that is not a random sample, the distributions we report describe these collections rather than their genres in full.
 
Two directions would address these limitations. A study with human annotators blind to the model labels would yield bias-corrected prevalence and, in particular, the external coverage of MER and ANT that no independent key in the present study reaches. A complementary design that codes the same arguments independently for their structural relation and for their proof method would test the grounding of the relations in the tropes that we have here taken as given, with alignment supporting that grounding and independence showing it to be decorative.
 
Beyond these questions of method, the flashes of intuition that allow researchers to see how an intractable problem might be solved remain largely resistant to systematic account. A framework such as ours can map the relations through which a proof is organized, but not the moment of insight that finds it. On that, obedient to the lemma inscribed over the entrance of Plato's Academy (``\gk{μηδεὶς ἀγεωμέτρητος εἰσίτω} -- Let no one ignorant of geometry enter here''), we can only stand modestly at the threshold.

\bibliographystyle{abbrvnat}
\bibliography{references}

@article{AppelHaken1976,
  author  = {Appel, Kenneth and Haken, Wolfgang},
  title   = {Every Planar Map is Four Colorable},
  journal = {Bulletin of the American Mathematical Society},
  volume  = {82},
  number  = {5},
  pages   = {711--712},
  year    = {1976},
  doi     = {10.1090/S0002-9904-1976-14122-5},
  }

@book{Burke1969,
  author    = {Burke, Kenneth},
  title     = {A Grammar of Motives},
  publisher = {University of California Press},
  address   = {Berkeley, CA, USA}, 
  year      = {1969}
}

@book{Casanova1987,
  author    = {Casanova, Marco Ant{\^o}nio and Giorno, Fernando Ant{\^o}nio de Castro and Furtado, Antonio Luz},
  title     = {Programa{\c c}{\~a}o em L{\'o}gica e a Linguagem {Prolog}},
  publisher = {Editora Edgard Bl{\"u}cher},
  address   = {S{\~a}o Paulo, Brazil},
  year      = {1987}
}

@inproceedings{Furtado2001,
  author    = {Furtado, Antonio L. and Ciarlini, Angelo E. M.},
  title     = {Constructing Libraries of Typical Plans},
  booktitle = {Advanced Information Systems Engineering: 13th International Conference, {CAiSE} 2001, Interlaken, Switzerland, June 4--8, 2001, Proceedings},
  editor    = {Dittrich, Klaus R. and Geppert, Andreas and Norrie, Moira C.},
  series    = {Lecture Notes in Computer Science},
  volume    = {2068},
  pages     = {124--139},
  year      = {2001},
  publisher = {Springer},
  address   = {Berlin, Heidelberg, Germany},
  doi       = {10.1007/3-540-45341-5_9}
}

@article{Ciarlini2009,
  author    = {Ciarlini, Angelo E. M. and Barbosa, Simone D. J. and Casanova, Marco A. and Furtado, Antonio L.},
  title     = {Event Relations in Plan-Based Plot Composition},
  journal   = {Computers in Entertainment},
  volume    = {7},
  number    = {4},
  articleno = {55},
  pages     = {1--37},
  numpages  = {37},
  year      = {2009},
  publisher = {Association for Computing Machinery},
  address   = {New York, NY, USA},
  doi       = {10.1145/1658866.1658874}
}

@book{Chandler2002,
  author    = {Chandler, Daniel},
  title     = {Semiotics: The Basics},
  series    = {The Basics},
  edition   = {1st},
  publisher = {Routledge},
  address   = {London, UK},
  year      = {2002}
}

@book{Chateaubriand2001,
  author    = {Chateaubriand, Oswaldo},
  title     = {Logical Forms. Part {I}: Truth and Description},
  series    = {Cole{\c c}{\~a}o CLE},
  volume    = {34},
  edition   = {1st},
  publisher = {Centro de L{\'o}gica, Epistemologia e Hist{\'o}ria da Ci{\^e}ncia (CLE), Universidade Estadual de Campinas},
  address   = {Campinas, SP, Brazil},
  pages     = {442},
  year      = {2001}
}

@book{Culler1981,
  author    = {Culler, Jonathan},
  title     = {The Pursuit of Signs: Semiotics, Literature, Deconstruction},
  edition   = {1st},
  publisher = {Routledge and Kegan Paul},
  address   = {London, UK},
  year      = {1981}
}

@book{Enderton1972,
  author    = {Enderton, Herbert B.},
  title     = {A Mathematical Introduction to Logic},
  edition   = {1st},
  publisher = {Academic Press},
  address   = {New York, NY, USA},
  year      = {1972}
}

@book{Euclid1956,
  author    = {Euclid},
  title     = {The Thirteen Books of the Elements},
  edition   = {2nd},
  publisher = {Dover Publications},
  address   = {New York, NY, USA},
  year      = {1956},
  note      = {Translated by Sir Thomas L. Heath}
}

@article{Faltings1995,
  author  = {Faltings, Gerd},
  title   = {The Proof of {Fermat}'s Last Theorem by {R. Taylor} and {A. Wiles}},
  journal = {Notices of the American Mathematical Society},
  volume  = {42},
  number  = {7},
  pages   = {743--746},
  year    = {1995},
  url     = {https://www.ams.org/notices/199507/faltings.pdf},
  note    = {Translated by Uwe F. Mayer}
}

@article{Furtado1992,
  author    = {Furtado, Antonio L.},
  title     = {Analogy by Generalization---and the Quest of the Grail},
  journal   = {ACM SIGPLAN Notices},
  volume    = {27},
  number    = {1},
  pages     = {105--113},
  year      = {1992},
  publisher = {Association for Computing Machinery},
  address   = {New York, NY, USA},
  doi       = {10.1145/130722.130741}
}

@book{Han2011,
  author    = {Han, Jiawei and Kamber, Micheline and Pei, Jian},
  title     = {Data Mining: Concepts and Techniques},
  series    = {The Morgan Kaufmann Series in Data Management Systems},
  edition   = {3rd},
  publisher = {Morgan Kaufmann},
  address   = {Waltham, MA, USA},
  pages     = {744},
  year      = {2011},
  doi       = {10.1016/C2009-0-61819-5}
}

@incollection{Jakobson1981,
  author    = {Jakobson, Roman},
  title     = {Linguistics and Poetics},
  booktitle = {Selected Writings. Volume {III}: Poetry of Grammar and Grammar of Poetry},
  editor    = {Rudy, Stephen},
  volume    = {3},
  pages     = {18--51},
  publisher = {Mouton},
  address   = {The Hague, Netherlands},
  year      = {1981},
}

@book{Jung1981,
  author    = {Jung, Carl Gustav},
  title     = {The Archetypes and the Collective Unconscious},
  series    = {The Collected Works of {C.~G.} Jung},
  volume    = {9},
  edition   = {2nd},
  publisher = {Princeton University Press},
  address   = {Princeton, NJ, USA},
  year      = {1981}
}

@incollection{Karp1972,
  author    = {Karp, Richard M.},
  title     = {Reducibility Among Combinatorial Problems},
  booktitle = {Complexity of Computer Computations},
  editor    = {Miller, Raymond E. and Thatcher, James W. and Bohlinger, Jean D.},
  series    = {The IBM Research Symposia Series},
  pages     = {85--103},
  publisher = {Plenum Press},
  address   = {New York, NY, USA},
  year      = {1972},
  doi       = {10.1007/978-1-4684-2001-2_9}
}

@book{Kolodner1993,
  author    = {Kolodner, Janet L.},
  title     = {Case-Based Reasoning},
  series    = {The Morgan Kaufmann Series in Representation and Reasoning},
  publisher = {Morgan Kaufmann},
  address   = {San Mateo, CA, USA},
  pages     = {668},
  year      = {1993}
}

@book{Lakoff2003,
  author    = {Lakoff, George and Johnson, Mark},
  title     = {Metaphors We Live By},
  publisher = {University of Chicago Press},
  address   = {Chicago, IL, USA},
  pages     = {276},
  year      = {2003},
  doi       = {10.7208/chicago/9780226470993.001.0001}
}

@book{Lakoff2000,
  author    = {Lakoff, George and N{\'u}{\~n}ez, Rafael E.},
  title     = {Where Mathematics Comes From: How the Embodied Mind Brings Mathematics into Being},
  edition   = {1st},
  publisher = {Basic Books},
  address   = {New York, NY, USA},
  pages     = {493},
  year      = {2000}
}

@book{Peirce1998,
  author    = {Peirce, Charles S.},
  title     = {The Essential Peirce: Selected Philosophical Writings, Volume 2 (1893--1913)},
  volume    = {2},
  publisher = {Indiana University Press},
  address   = {Bloomington, IN, USA},
  year      = {1998},
  note      = {Edited by the Peirce Edition Project}
}

@book{Plato1926,
  author    = {Plato},
  title     = {Cratylus. Parmenides. Greater Hippias. Lesser Hippias},
  series    = {Loeb Classical Library},
  number    = {167},
  publisher = {Harvard University Press},
  address   = {Cambridge, MA, USA},
  year      = {1926},
  note      = {Translated by Harold North Fowler}
}

@book{Quintilian2001,
  author    = {Quintilian},
  title     = {The Orator's Education, Volume {III}: Books 6--8},
  series    = {Loeb Classical Library},
  number    = {126},
  publisher = {Harvard University Press},
  address   = {Cambridge, MA, USA},
  year      = {2001},
  note      = {Edited and translated by Donald A. Russell.}
}

@book{Ramus2010,
  author    = {Ramus, Peter},
  title     = {Arguments in Rhetoric Against {Quintilian}: Translation and Text of
               {Peter Ramus}'s {Rhetoricae Distinctiones in Quintilianum} (1549)},
  series    = {Landmarks in Rhetoric and Public Address},
  publisher = {Southern Illinois University Press},
  address   = {Carbondale, IL, USA},
  pages     = {234},
  year      = {2010},
  note      = {Edited by James J. Murphy; translated by Carole Newlands.}
}

@book{Saussure1995,
  author    = {de Saussure, Ferdinand},
  title     = {Cours de linguistique g{\'e}n{\'e}rale},
  series    = {Grande Biblioth{\`e}que Payot},
  publisher = {Payot},
  address   = {Paris, France},
  year      = {1995},
  note      = {Edited by Charles Bally, Albert Sechehaye and Albert Riedlinger.}
}

@book{Schank1977,
  author    = {Schank, Roger C. and Abelson, Robert P.},
  title     = {Scripts, Plans, Goals and Understanding: An Inquiry into Human Knowledge Structures},
  series    = {The Artificial Intelligence Series},
  publisher = {Lawrence Erlbaum Associates},
  address   = {Hillsdale, NJ, USA},
  pages     = {248},
  year      = {1977}
}

@book{Toulmin2003,
  author    = {Toulmin, Stephen E.},
  title     = {The Uses of Argument},
  edition   = {Updated},
  publisher = {Cambridge University Press},
  address   = {Cambridge, UK},
  pages     = {262},
  year      = {2003},
  doi       = {10.1017/CBO9780511840005}
}

@article{Turner1998,
  author  = {Fauconnier, Gilles and Turner, Mark},
  title   = {Conceptual Integration Networks},
  journal = {Cognitive Science},
  volume  = {22},
  number  = {2},
  pages   = {133--187},
  year    = {1998},
  doi     = {10.1207/s15516709cog2202_1}
}

@book{Vico1968,
  author    = {Vico, Giambattista},
  title     = {The New Science of {Giambattista Vico}: Unabridged Translation
               of the Third Edition (1744)},
  publisher = {Cornell University Press},
  address   = {Ithaca, NY, USA},
  year      = {1968}
}

@incollection{Webber1986,
  author    = {Webber, Bonnie L.},
  title     = {Questions, Answers and Responses: Interacting with Knowledge-Base Systems},
  booktitle = {On Knowledge Base Management Systems: Integrating Artificial Intelligence
               and Database Technologies},
  editor    = {Brodie, Michael L. and Mylopoulos, John},
  series    = {Topics in Information Systems},
  pages     = {365--401},
  publisher = {Springer-Verlag},
  address   = {New York, NY, USA},
  year      = {1986},
  doi       = {10.1007/978-1-4612-4980-1_30}
}

@book{White1973,
  author    = {White, Hayden},
  title     = {Metahistory: The Historical Imagination in Nineteenth-Century Europe},
  publisher = {Johns Hopkins University Press},
  address   = {Baltimore, MD, USA},
  year      = {1973}
}

@article{Wiles1995,
  author  = {Wiles, Andrew},
  title   = {Modular Elliptic Curves and {Fermat}'s Last Theorem},
  journal = {Annals of Mathematics},
  series  = {Second Series},
  volume  = {141},
  number  = {3},
  pages   = {443--551},
  year    = {1995},
  doi     = {10.2307/2118559}
}

@article{Winston1987,
  author  = {Winston, Morton E. and Chaffin, Roger and Herrmann, Douglas},
  title   = {A Taxonomy of Part-Whole Relations},
  journal = {Cognitive Science},
  volume  = {11},
  number  = {4},
  pages   = {417--444},
  year    = {1987},
  doi     = {10.1207/s15516709cog1104_2}
}

@inproceedings{Welleck2021,
  author    = {Welleck, Sean and Liu, Jiacheng and Le Bras, Ronan and
               Hajishirzi, Hannaneh and Choi, Yejin and Cho, Kyunghyun},
  title     = {{NaturalProofs}: Mathematical Theorem Proving in Natural Language},
  booktitle = {Proceedings of the Neural Information Processing Systems Track on
               Datasets and Benchmarks 1 ({NeurIPS} Datasets and Benchmarks 2021), Round 1},
  year      = {2021},
  url       = {https://datasets-benchmarks-proceedings.neurips.cc/paper/2021/hash/d9d4f495e875a2e075a1a4a6e1b9770f-Abstract-round1.html}
}

@inproceedings{PeldszusStede2016,
  author    = {Peldszus, Andreas and Stede, Manfred},
  title     = {An Annotated Corpus of Argumentative Microtexts},
  booktitle = {Argumentation and Reasoned Action: Proceedings of the 1st European
               Conference on Argumentation, Lisbon 2015},
  editor    = {Mohammed, Dima and Lewi{\'n}ski, Marcin},
  volume    = {2},
  pages     = {801--815},
  publisher = {College Publications},
  address   = {London, UK},
  year      = {2016}
}

@article{Habernal2024,
  author  = {Habernal, Ivan and Faber, Daniel and Recchia, Nicola and
             Bretthauer, Sebastian and Gurevych, Iryna and
             {Spiecker genannt D{\"o}hmann}, Indra and Burchard, Christoph},
  title   = {Mining Legal Arguments in Court Decisions},
  journal = {Artificial Intelligence and Law},
  volume  = {32},
  number  = {3},
  pages   = {557--594},
  year    = {2024},
  doi     = {10.1007/s10506-023-09361-y}
}

@book{Krippendorff2004,
  author    = {Krippendorff, Klaus},
  title     = {Content Analysis: An Introduction to Its Methodology},
  edition   = {2nd},
  publisher = {Sage Publications},
  address   = {Thousand Oaks, CA, USA},
  pages     = {xxiii, 413},
  year      = {2004}
}

@article{Newcombe1998,
  author  = {Newcombe, Robert G.},
  title   = {Interval Estimation for the Difference between Independent Proportions:
             Comparison of Eleven Methods},
  journal = {Statistics in Medicine},
  volume  = {17},
  number  = {8},
  pages   = {873--890},
  month   = apr,
  year    = {1998},
  doi     = {10.1002/(SICI)1097-0258(19980430)17:8<873::AID-SIM779>3.0.CO;2-I}
}

@book{Walton1996,
  author    = {Walton, Douglas N.},
  title     = {Argumentation Schemes for Presumptive Reasoning},
  publisher = {Lawrence Erlbaum Associates},
  address   = {Mahwah, NJ, USA},
  year      = {1996},
  doi       = {10.4324/9780203811160}
}

@book{WaltonReedMacagno2008,
  author    = {Walton, Douglas N. and Reed, Christopher and Macagno, Fabrizio},
  title     = {Argumentation Schemes},
  publisher = {Cambridge University Press},
  address   = {Cambridge, UK},
  pages     = {443},
  year      = {2008},
  doi       = {10.1017/CBO9780511802034}
}

@inproceedings{Lima2023,
  author    = {{de Lima}, Edirlei Soares and Casanova, Marco A. and
               Feij{\'o}, Bruno and Furtado, Antonio L.},
  title     = {Semiotic Structuring in Movie Narrative Generation},
  booktitle = {Entertainment Computing -- {ICEC} 2023},
  editor    = {Ciancarini, Paolo and Di Iorio, Angelo and Hlavacs, Helmut and
               Poggi, Francesca},
  series    = {Lecture Notes in Computer Science},
  volume    = {14455},
  pages     = {161--175},
  publisher = {Springer Nature Singapore},
  address   = {Singapore},
  year      = {2023},
  doi       = {10.1007/978-981-99-8248-6_13}
}

@misc{Lima2026frye,
  author        = {{de Lima}, Edirlei Soares and Casanova, Marco A. and
                   Furtado, Antonio L.},
  title         = {Revisiting {Northrop Frye}'s Four Myths Theory with
                   Large Language Models},
  year          = {2026},
  eprint        = {2602.15678},
  archivePrefix = {arXiv},
  primaryClass  = {cs.CL},
  doi           = {10.48550/arXiv.2602.15678},
  howpublished  = {arXiv:2602.15678 [cs.CL]},
  url           = {https://doi.org/10.48550/arXiv.2602.15678}
}

@article{MannThompson1988,
  author  = {Mann, William C. and Thompson, Sandra A.},
  title   = {Rhetorical Structure Theory: Toward a Functional Theory of Text Organization},
  journal = {Text: Interdisciplinary Journal for the Study of Discourse},
  volume  = {8},
  number  = {3},
  pages   = {243--281},
  year    = {1988},
  doi     = {10.1515/text.1.1988.8.3.243}
}

@inproceedings{Prasad2008,
  author    = {Prasad, Rashmi and Dinesh, Nikhil and Lee, Alan and
               Miltsakaki, Eleni and Robaldo, Livio and Joshi, Aravind and
               Webber, Bonnie L.},
  title     = {The {Penn Discourse TreeBank} 2.0},
  booktitle = {Proceedings of the Sixth International Conference on Language
               Resources and Evaluation ({LREC}'08)},
  editor    = {Calzolari, Nicoletta and Choukri, Khalid and Maegaard, Bente and
               Mariani, Joseph and Odijk, Jan and Piperidis, Stelios and Tapias, Daniel},
  pages     = {2961--2968},
  year      = {2008},
  publisher = {European Language Resources Association (ELRA)},
  address   = {Marrakech, Morocco},
  url       = {https://aclanthology.org/L08-1093/}
}

@misc{Tornberg2023,
  author        = {T{\"o}rnberg, Petter},
  title         = {{ChatGPT-4} Outperforms Experts and Crowd Workers in Annotating
                   Political {Twitter} Messages with Zero-Shot Learning},
  year          = {2023},
  eprint        = {2304.06588},
  archivePrefix = {arXiv},
  primaryClass  = {cs.CL},
  doi           = {10.48550/arXiv.2304.06588},
  howpublished  = {arXiv:2304.06588 [cs.CL]},
  url           = {https://doi.org/10.48550/arXiv.2304.06588}
}

@inproceedings{Calderon2025,
  author    = {Calderon, Nitay and Reichart, Roi and Dror, Rotem},
  title     = {The Alternative Annotator Test for {LLM}-as-a-Judge: How to Statistically
               Justify Replacing Human Annotators with {LLMs}},
  booktitle = {Proceedings of the 63rd Annual Meeting of the Association for
               Computational Linguistics (Volume 1: Long Papers)},
  editor    = {Che, Wanxiang and Nabende, Joyce and Shutova, Ekaterina and
               Pilehvar, Mohammad Taher},
  pages     = {16051--16081},
  year      = {2025},
  publisher = {Association for Computational Linguistics},
  address   = {Vienna, Austria},
  doi       = {10.18653/v1/2025.acl-long.782}
}

@article{Lima2016,
  author  = {{de Lima}, Edirlei Soares and Feij{\'o}, Bruno and
             Casanova, Marco A. and Furtado, Antonio L.},
  title   = {Storytelling Variants Based on Semiotic Relations},
  journal = {Entertainment Computing},
  volume  = {17},
  pages   = {31--44},
  year    = {2016},
  doi     = {10.1016/j.entcom.2016.08.003}
}

@article{vanBlerck2025,
  author  = {{van Blerck}, Irene C. E. and {de Lima}, Edirlei Soares and
             Neggers, Margot M. E. and Calders, Toon},
  title   = {Unveiling Gender Bias in {LLM}-Generated Hero and Heroine Narratives},
  journal = {Entertainment Computing},
  volume  = {55},
  pages   = {100972},
  year    = {2025},
  doi     = {10.1016/j.entcom.2025.100972}
}

@inproceedings{Li2024,
  author    = {Li, Zhaoyu and Sun, Jialiang and Murphy, Logan and Su, Qidong and
               Li, Zenan and Zhang, Xian and Yang, Kaiyu and Si, Xujie},
  title     = {A Survey on Deep Learning for Theorem Proving},
  booktitle = {First Conference on Language Modeling ({COLM})},
  year      = {2024},
  url       = {https://openreview.net/forum?id=zlw6AHwukB}
}

@article{Webb2023,
  author  = {Webb, Taylor and Holyoak, Keith J. and Lu, Hongjing},
  title   = {Emergent Analogical Reasoning in Large Language Models},
  journal = {Nature Human Behaviour},
  volume  = {7},
  number  = {9},
  pages   = {1526--1541},
  year    = {2023},
  doi     = {10.1038/s41562-023-01659-w}
}

@article{LewisMitchell2025,
  author  = {Lewis, Martha and Mitchell, Melanie},
  title   = {Evaluating the Robustness of Analogical Reasoning in Large Language Models},
  journal = {Transactions on Machine Learning Research},
  year    = {2025},
  url     = {https://openreview.net/forum?id=t5cy5v9wph},
}

@book{Polya1945,
  author    = {P{\'o}lya, George},
  title     = {How to Solve It: A New Aspect of Mathematical Method},
  edition   = {1st},
  publisher = {Princeton University Press},
  address   = {Princeton, NJ, USA},
  pages     = {204},
  year      = {1945}
}

@book{Polya1954,
  author    = {P{\'o}lya, George},
  title     = {Mathematics and Plausible Reasoning. Volume {I}:
               Induction and Analogy in Mathematics},
  volume    = {1},
  publisher = {Princeton University Press},
  address   = {Princeton, NJ, USA},
  year      = {1954}
}

@book{Polya2014,
  author    = {P{\'o}lya, George},
  title     = {Mathematics and Plausible Reasoning [Two Volumes in One]},
  publisher = {Martino Fine Books},
  address   = {Eastford, CT, USA},
  year      = {2014},
  pages     = {498}
}

@book{Lakatos1976,
  author    = {Lakatos, Imre},
  title     = {Proofs and Refutations: The Logic of Mathematical Discovery},
  publisher = {Cambridge University Press},
  address   = {Cambridge, UK},
  year      = {1976},
  note      = {Edited by John Worrall and Elie Zahar}
}

@article{Lima2026,
  author  = {{de Lima}, Edirlei Soares and Neggers, Margot M. E. and
             Casanova, Marco A. and Feij{\'o}, Bruno and Furtado, Antonio L.},
  title   = {Trait-Guided Detective Story Generation in Seven Classic
             Investigative Styles with Large Language Models},
  journal = {Entertainment Computing},
  volume  = {58},
  pages   = {101166},
  year    = {2026},
  doi     = {10.1016/j.entcom.2026.101166}
}

@misc{Lima2025,
  author        = {{de Lima}, Edirlei Soares and Casanova, Marco A. and
                   Feij{\'o}, Bruno and Furtado, Antonio L.},
  title         = {Characterizing the Investigative Methods of Fictional Detectives
                   with Large Language Models},
  year          = {2025},
  eprint        = {2505.07601},
  archivePrefix = {arXiv},
  primaryClass  = {cs.CL},
  doi           = {10.48550/arXiv.2505.07601},
  howpublished  = {arXiv:2505.07601 [cs.CL]},
  url           = {https://doi.org/10.48550/arXiv.2505.07601}
}

@inproceedings{Lima2023sbgames,
  author    = {{de Lima}, Edirlei Soares and Feij{\'o}, Bruno and
               Casanova, Marco A. and Furtado, Antonio L.},
  title     = {{ChatGeppetto} -- An {AI}-Powered Storyteller},
  booktitle = {Proceedings of the 22nd Brazilian Symposium on Games and
               Digital Entertainment ({SBGames} '23)},
  series    = {ACM International Conference Proceeding Series},
  pages     = {28--37},
  month     = nov,
  year      = {2023},
  publisher = {Association for Computing Machinery},
  address   = {New York, NY, USA},
  doi       = {10.1145/3631085.3631302}
}

@article{Lima2025b,
  author  = {{de Lima}, Edirlei Soares and Neggers, Margot M. E. and
             Feij{\'o}, Bruno and Casanova, Marco A. and Furtado, Antonio L.},
  title   = {An {AI}-Powered Approach to the Semiotic Reconstruction of Narratives},
  journal = {Entertainment Computing},
  volume  = {52},
  pages   = {100810},
  year    = {2025},
  doi     = {10.1016/j.entcom.2024.100810}
}

@book{Ranganathan1967,
  author    = {Ranganathan, Shiyali Ramamrita},
  title     = {Ramanujan: The Man and the Mathematician},
  series    = {Great Thinkers of India Series},
  volume    = {1},
  publisher = {Asia Publishing House},
  address   = {Bombay, India},
  year      = {1967}
}

@inproceedings{deMoura2021,
  author    = {de Moura, Leonardo and Ullrich, Sebastian},
  title     = {The {Lean} 4 Theorem Prover and Programming Language},
  booktitle = {Automated Deduction -- {CADE} 28: 28th International Conference on
               Automated Deduction, Virtual Event, July 12--15, 2021, Proceedings},
  editor    = {Platzer, Andr{\'e} and Sutcliffe, Geoff},
  series    = {Lecture Notes in Computer Science},
  volume    = {12699},
  pages     = {625--635},
  year      = {2021},
  publisher = {Springer},
  address   = {Cham, Switzerland},
  doi       = {10.1007/978-3-030-79876-5_37}
}

@inproceedings{Tzachristas2026,
  author    = {Tzachristas, Ioannis and Tzachristas, Georgios and Sui, Aifen},
  title     = {Open Problems Solved by {LLMs}? {A} Survey of Verifiable
               Mathematical Discovery},
  booktitle = {Proceedings of The Big Picture v2: Crafting a Research Narrative},
  pages     = {10--21},
  year      = {2026},
  publisher = {Association for Computational Linguistics},
  address   = {San Diego, CA, USA},
  url       = {https://aclanthology.org/2026.bigpicture-main.2/}
}

@misc{Ke2026,
  author        = {Ke, Yisi and Huang, Tianyu and Shu, Yankai and He, Di and
                   Gai, Jingchu and Wang, Liwei},
  title         = {Towards Solving the {Gilbert--Pollak} Conjecture via
                   Large Language Models},
  year          = {2026},
  eprint        = {2601.22365},
  archivePrefix = {arXiv},
  doi           = {10.48550/arXiv.2601.22365},
  howpublished  = {2601.22365 [cs.DM]},
  url           = {https://doi.org/10.48550/arXiv.2601.22365}
}

@inproceedings{Welleck2022,
  author    = {Welleck, Sean and Liu, Jiacheng and Lu, Ximing and
               Hajishirzi, Hannaneh and Choi, Yejin},
  title     = {{NaturalProver}: Grounded Mathematical Proof Generation with
               Language Models},
  booktitle = {Advances in Neural Information Processing Systems 35
               ({NeurIPS} 2022)},
  year      = {2022},
  publisher = {Curran Associates, Inc.},
  url       = {https://openreview.net/forum?id=rhdfTOiXBng}
}

@misc{Polu2020,
  author        = {Polu, Stanislas and Sutskever, Ilya},
  title         = {Generative Language Modeling for Automated Theorem Proving},
  year          = {2020},
  eprint        = {2009.03393},
  archivePrefix = {arXiv},
  doi           = {10.48550/arXiv.2009.03393},
  howpublished  = {2009.03393 [cs.LG]},
  url           = {https://doi.org/10.48550/arXiv.2009.03393}
}

\newpage
\begin{appendices}

\section{The Annotation Prompt}
\label{app:prompt}

This appendix reproduces the complete system prompt used in the annotation run described in Section~\ref{sec:eval:instrument}, exactly as supplied to the models. The four relation definitions are shown here in a fixed order (SYN, PAR, ANT, MER). At run time they are presented in a randomized order per call, with NONE pinned last, to prevent position in the list from influencing the label. The passage to be classified is supplied in a separate user message consisting of the word \texttt{Passage:} followed by the passage text.

SYSTEM PROMPT:

\begin{lstlisting}[basicstyle=\ttfamily\footnotesize,breaklines=true,columns=fullflexible,literate={¬}{{$\neg$}}1]
Classify how a passage of reasoning establishes its claim.

Each passage argues for some claim. Call the claim S, and call whatever the passage rests that claim on S*. Your job is to identify the relationship between S and S*. That is, what KIND of move the passage makes to get from its support to its conclusion.

The categories:
- SYN (syntagmatic; contiguity, sequence; trope: metonymy; method: inference): S is a logical consequence of S*. The reasoner locates a rule or prior result and applies it to reach S. Includes syllogism, rule application, chained deduction, and defeasible warrant-based argument.
- PAR (paradigmatic; similarity, alternatives; trope: metaphor; method: analogy): After suitable mappings, the relevant features of S are converted into features of S*. The reasoner solves the mapped problem instead of the original one. The analogue often lies in another domain, as when a scheduling problem is mapped onto graph colouring, but it need not: mapping one case onto a structurally similar case within the same domain is still PAR, as when a problem is solved by blending a situation with a parallel version of itself. What matters is that a correspondence is drawn between two cases and the argument runs through it. Includes reduction, structural analogy, argument from a parallel case, and conceptual blending.
- ANT (antithetic; opposition, negation; trope: irony; method: contradiction): S* is shown to be inconsistent under the assumption of ¬S. Includes reductio ad absurdum and any argument whose force comes from the untenability of the denial. Concession is not ANT. Granting a point and then arguing past it ("Of course X, but Y"; "Admittedly X; nevertheless Y") is a rhetorical move, not an argument from inconsistency; label such passages by whatever establishes the main claim. Nor is mere contrast between two things ANT. The test is whether denying the claim is shown to lead to something untenable.
- MER (meronymic; hierarchy, details; trope: synecdoche; method: case analysis): S* is a set of statements into which S decomposes exhaustively, each handled separately. Includes case splits, exhaustion, and finite induction (base case + induction step).
- NONE: the artifact establishes S by a move none of the four captures, or makes no identifiable argumentative move at all. Do not stretch a category to avoid NONE; its rate is a result, not a failure. But NONE is not a way to record uncertainty. If the passage does make one of the four moves and you are merely unsure which, pick the best fit and set confidence to 1. Reserve NONE for passages where no category applies, for instance a passage that asserts a conclusion without supporting it, or recites facts without drawing an inference. Note that a practical or normative conclusion drawn from stated circumstances is still SYN: the warrant may be implicit and defeasible without ceasing to be an inference.

How to decide:

- Judge only what is in the passage. Do not use outside knowledge of the topic, and do not reconstruct an argument the text does not actually make.
- Passages may be excerpts, especially from longer documents. Some will begin or end mid-argument. Classify the move the excerpt itself makes; do not guess at what surrounded it.
- The dominant category is the one the argument depends on: remove that move and the passage no longer establishes its claim. There is exactly one.
- Subordinate categories are for moves doing real supporting work. Leave the list empty when there are none. Do not add a category as subordinate merely because the passage contains some faint trace of it; if that field fires on everything it carries no information.
- Choose NONE when the passage states a conclusion without arguing for it, recites facts or rules without drawing an inference, or makes a move none of the categories fits. NONE is a legitimate answer and you should expect to use it. Do not stretch a category to avoid it.
- These categories are not equally common. Do not try to balance your answers across a set of passages, and do not assume any category must appear.

Work in this order, then respond with JSON only, no other text:

1. claim: the claim the passage is establishing (S), in your own words, one sentence.
2. support: what the passage rests that claim on (S*), one sentence.
3. reasoning: what kind of move connects support to claim, one sentence.
4. dominant, subordinate, confidence, as defined above.

{"claim": "...", "support": "...", "reasoning": "...", "dominant": "...", "subordinate": [...], "confidence": N}

confidence: 1 if you are guessing, 2 if plausible, 3 if clear.
\end{lstlisting}

\end{appendices} 

\end{document}